\pdfoutput=1

\documentclass[11pt]{article}

\usepackage[final]{coling}

\usepackage{times}
\usepackage{latexsym}

\usepackage[T1]{fontenc}

\usepackage[utf8]{inputenc}

\usepackage{microtype}

\usepackage{inconsolata}

\usepackage{graphicx}
\usepackage{xcolor}         
\usepackage{multirow, makecell}
\usepackage{enumitem}
\usepackage{wrapfig,lipsum,booktabs}
\usepackage{diagbox} 
\newcommand{\eg}{\textit{e.g., }}
\newcommand{\ie}{\textit{i.e., }}

\title{Graph Reasoning with Large Language Models via Pseudo-code Prompting}

\author{
 \textbf{Konstantinos Skianis\textsuperscript{1}},
 \textbf{Giannis Nikolentzos\textsuperscript{2}},
 \textbf{Michalis Vazirgiannis\textsuperscript{3}}
\\
 \textsuperscript{1}University of Ioannina, Greece \\
 \textsuperscript{2}University of Peloponnese, Greece \\
 \textsuperscript{3}LIX, \'{E}cole Polytechnique, IP Paris,
France 
\\
 \small{
   \textbf{Corresponding author:} \href{mailto:kskianis@cse.uoi.gr}{kskianis@cse.uoi.gr}
 }
}

\begin{document}
\maketitle
\begin{abstract}
Large language models (LLMs) have recently achieved remarkable success in various reasoning tasks in the field of natural language processing.
This success of LLMs has also motivated their use in graph-related tasks.
Among others, recent work has explored whether LLMs can solve graph problems such as counting the number of connected components of a graph or computing the shortest path distance between two nodes.
Although LLMs possess preliminary graph reasoning abilities, they might still struggle to solve some seemingly simple problems.
In this paper, we investigate whether prompting via pseudo-code instructions can improve the performance of LLMs in solving graph problems.
This approach not only aligns the model's reasoning with algorithmic logic but also imposes a structured, modular approach to problem-solving that is inherently transparent and interpretable.
Our experiments demonstrate that using pseudo-code instructions generally improves the performance of all considered LLMs.
The graphs, pseudo-code prompts, and evaluation code are publicly available\footnote{\url{https://github.com/y3nk0/graph-reasoning-llms}}.
\end{abstract}

\section{Introduction}

Recently, the artificial intelligence community has witnessed great advancements in the field of large language models (LLMs)~\cite{devlin2019bert,brown2020language,ouyang2022training}.
Those models have captured intense public and academic interest, while the success of LLMs in different domains such as in medicine~\cite{thirunavukarasu2023large} and in software engineering~\cite{poesia2022synchromesh} has boosted hopes that these models could potentially pave the way for the development of Artificial General Intelligence~\cite{bubeck2023sparks}.
These advancements have been made possible not only due to breakthroughs in the field of machine learning, such as the introduction of the Transformer~\cite{vaswani2017attention}, but also due to the availability of massive amounts of data and the increase of computational power.

While LLMs were originally designed for textual data, they have already been utilized in settings that go beyond their initial application context.
In several of those settings, a graph structure is explicitly or implicitly involved.
For example, in world modeling, LLMs are commonly employed to generate knowledge graphs for text games in order to improve an agent's ability to efficiently operate in complex environments~\cite{ammanabrolu2021learning}.
However, LLMs rely on unstructured text, and in those settings, they might fail to properly encode the different entities and their relationships.
This might lead to different issues, e.g. the models might fail to deduce some logical entailments or they might hallucinate, i.e. generate plausible-sounding responses that are factually incorrect.

Despite the preliminary success of LLMs in the aforementioned settings, it is still not entirely clear whether those models exhibit fundamental limitations that might constrain their applicability in those domains.
Some recent studies shed some light on this issue by investigating whether LLMs can actually reason with graphs~\cite{wang2023can,fatemi2024talk}.
In fact, those studies investigated whether LLMs can solve graph problems fed to them as natural language prompts.
The two studies employed different LLMs, and the reported results are somewhat ambivalent.
While in one study, it was shown that LLMs possess preliminary graph reasoning abilities~\cite{wang2023can}, in the other study, LLMs failed to solve basic graph tasks (\eg count the number of edges of a graph).

\begin{figure*}[t]
    \centering
\includegraphics[width=\textwidth,height=\textheight,keepaspectratio]{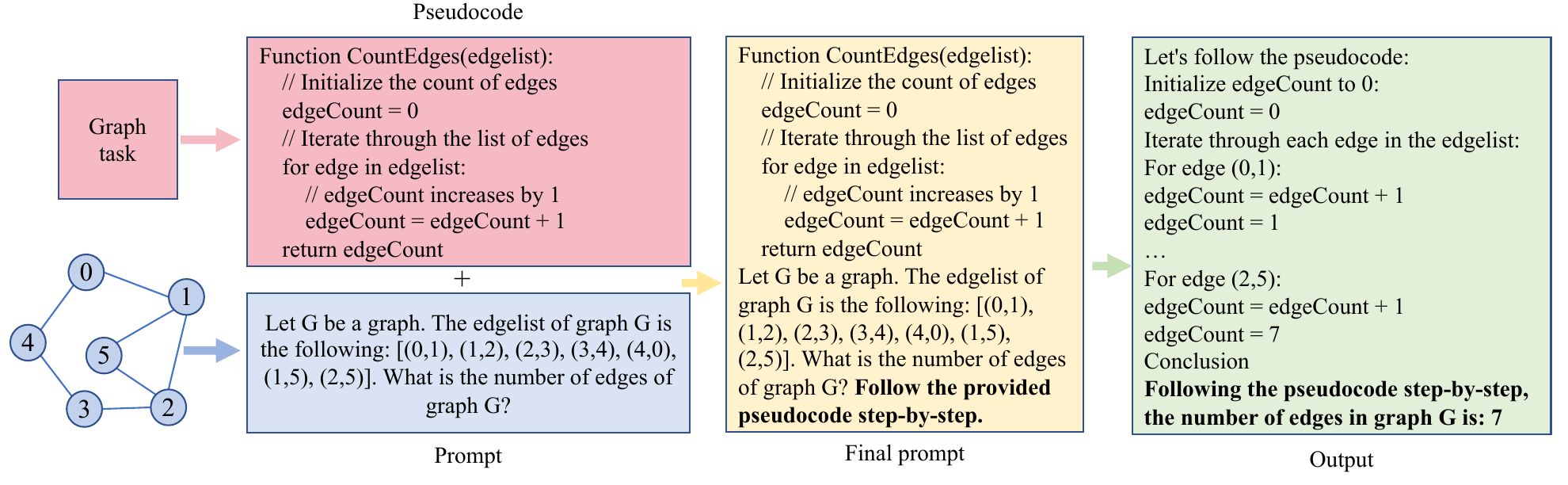}
    \caption{An illustration of our proposed method for graph reasoning using pseudo-code instructions.}
    \label{fig:proposed_method}
\end{figure*}
In this paper, we study whether prompt engineering can help us improve the performance of LLMs in solving graph algorithm problems.
Natural language instructions can be ambiguous and under-specified, and this might prevent models from returning the most accurate answer possible.
Furthermore, very detailed instructions might increase the complexity of reasoning and harm the model's performance.
Therefore, prompt engineering can significantly contribute to enhancing the capabilities of pre-trained LLMs.~\cite{liu2023pre}.
Different prompting strategies have been developed so far.
The idea to use prompts that encourage multi-step reasoning led to very successful methods such as the chain-of-thought (CoT) reasoning in few-shot settings~\cite{wei2022chain}, while it was also shown that LLMs can become decent zero-shot reasoners by just adding the prompt ``Let's think step by step''~\cite{kojima2022large}.
Here, we investigate whether the use of pseudo-code instructions for prompting can enhance the performance of LLMs in solving graph algorithm problems.
Pseudo-code can reduce the ambiguity present in natural language, but it also provides explicit and clear instructions on how to solve a problem.
An example of the proposed approach for prompting with pseudo-code is illustrated in Figure~\ref{fig:proposed_method}.
We study performance in $10$ graph reasoning tasks on two LLM families (GPT and Mixtral).
The obtained results indicate that the proposed method improves the performance of LLM mainly in tasks that they struggle to solve.

In summary, our paper makes the following contributions: 
\begin{itemize}[noitemsep, topsep=0pt]
    \item We release a new benchmark dataset of pseudo-code prompts for different graph problems to test the reasoning abilities of LLMs.
    \item We study the impact of these prompts on the performance of three LLMs in $10$ graph reasoning tasks.
    \item The experimental results demonstrate that augmenting prompts with pseudo-code can be useful for solving both simple, but also complex graph reasoning tasks.
\end{itemize}

\section{Related Work}

\paragraph{Large Language Models and graphs.}
Graph neural networks (GNNs) have been established as the standard neural architecture for performing machine learning on graphs since these models are invariant to permutations of the nodes of the input graph~\cite{zhou2020graph}.
Other common architectures, such as the family of recurrent neural networks, do not enjoy this property.
However, permutation-sensitive models such as the Transformer architecture~\cite{vaswani2017attention} can also deal with graph learning problems.
For example, it was shown in~\cite{kim2022pure} that if we treat both nodes and edges as independent tokens, augment them with token embeddings, and feed them to a Transformer, we obtain a powerful graph learner.
Some node classification datasets where the nodes are annotated with textual content have been treated as text classification datasets by ignoring the graph structure, and LLMs have been leveraged to classify the textual content~\cite{chen2024exploring}.
It was found that LLMs achieve good zero-shot performance on certain datasets.
Similar conclusions were also reached by other works~\cite{hu2023beyond}.
Real-world data is noisy and this also applies to graphs.
Thus, some works have leveraged LLMs to refine graphs~\cite{sun2023large,guo2024graphedit}.
In the GraphEdit method, the LLM is responsible for identifying noisy connections between irrelevant nodes and for discovering implicit dependencies between nodes based on the textual data associated with them~\cite{guo2024graphedit}.
Several works have investigated the potential of LLMs to enhance the performance of GNNs on text-attributed graphs~\cite{duan2023simteg,chen2024exploring,he2024harnessing}.
For instance, TAPE uses an LLM to extract predictions and explanations from the input text which serve as supplementary features for the downstream GNN model~\cite{he2024harnessing}.
The works closest to ours in this domain are the ones reported in~\cite{wang2023can} and in~\cite{fatemi2024talk}, which investigate whether LLMs can solve graph algorithm problems in natural language.
In this paper, we go one step further and study whether prompting with pseudo-code instructions can help LLMs better understand how to solve graph problems.

Not only LLMs have emerged as useful tools in graph learning tasks, but it turns out that the opposite is also true, \ie graphs can enhance LLMs~\cite{pan2024unifying}.
Even though LLMs have achieved great success in the past years, they still might suffer from different problems such as hallucinations, reduced factuality awareness, and limited explainability.
Knowledge graphs can help LLMs deal with those issues since they store extensive high-quality and reliable factual knowledge.
Therefore, to mitigate the aforementioned issues, knowledge graphs have been recently incorporated to improve the reasoning ability of LLMs~\cite{guan2024mitigating,luo2024reasoning}.

\paragraph{Prompt engineering.}
Prompt engineering seeks for the best way to describe a task such that an LLM can solve the task using its autoregressive token-based mechanism for generating text.
Prompt engineering is a resource-efficient approach in the sense that it does not require access to the internals of the model (\eg its parameters).
We can thus provide the model with a task description and ask it to solve the task even if it has never been trained on it.
Few-shot prompting aims to teach the language model how to solve a task by providing it with a small number of example tasks with solutions~\cite{brown2020language}. 
The model then learns from these examples and can solve similar tasks.
Chain-of-Thought (CoT) is a prompting technique, in which one includes a series of intermediate natural language reasoning steps that lead to the desired output~\cite{wei2022chain}.
CoT was shown to significantly improve the capability of LLMs to solve problems.
Zero-shot-CoT, another approach for prompting, simply adds the prompt ``\textit{Let's think step by step}'' before each answer to facilitate step-by-step thinking~\cite{kojima2022large}.
Zero-shot-CoT turned out to be the strongest zero-shot baseline, while LLMs were shown to be decent zero-shot reasoners.
However, Zero-shot-CoT might fail in some cases because of missing reasoning steps.
Prompting via pseudo-code instructions has also been recently explored for solving natural language processing tasks~\cite{mishra2023prompting}.
Program-of-thoughts prompting generates code to solve a task~\cite{chen2023program}.
It uses Python code to describe reasoning steps, and the computation is accomplished by a Python interpreter.
To improve the LLMs reasoning ability, some works have employed multiple rounds of prompting~\cite{jung2022maieutic}.
For instance, least-to-most prompting teaches language models how to solve a complex problem by decomposing
it into a series of simpler subproblems which are solved one after the other~\cite{zhou2023least}.
Self-Consistency is a scheme where multiple CoTs are generated and one of them is finally chosen~\cite{wang2023self}.
Tree of Thoughts (ToT)~\cite{yao2023tree} and Graph of Thoughts (GoT)~\cite{besta2024graph} are two schemes that model the LLM reasoning process with a tree and a graph, respectively.

When LLMs are leveraged to solve graph tasks, different graph encoding schemes can be utilized to transform graph-structured data into text (\eg list of edges, adjacency matrix, graph description language, etc.).
It was recently shown that input design indeed has a significant impact on the final result~\cite{guo2023gpt4graph}.
GraphText constructs a graph-syntax tree from the input graph, and then, the traversal of the graph-syntax tree leads to a prompt in natural language which can be fed to the LLM to perform graph reasoning.~\cite{zhao2023graphtext}.
More recently, continuous graph representations have been explored~\cite{perozzi2024let}.
The graph is mapped into a continuous vector via a GNN and this vector serves as input for the LLM.

\section{Proposed Methodology}
To investigate whether prompting with pseudo-code instructions can improve the capability of language models in reasoning with graphs, we focus on a wide range of graph tasks, we construct instances of those tasks and present them along with the pseudo-code that solves them as natural language queries to the language models.
We next give more details about the different graph tasks we consider in this paper and how the different problem instances are generated.

\begin{figure*}[t]
    \centering
\includegraphics[width=0.9\textwidth,height=\textheight,keepaspectratio]{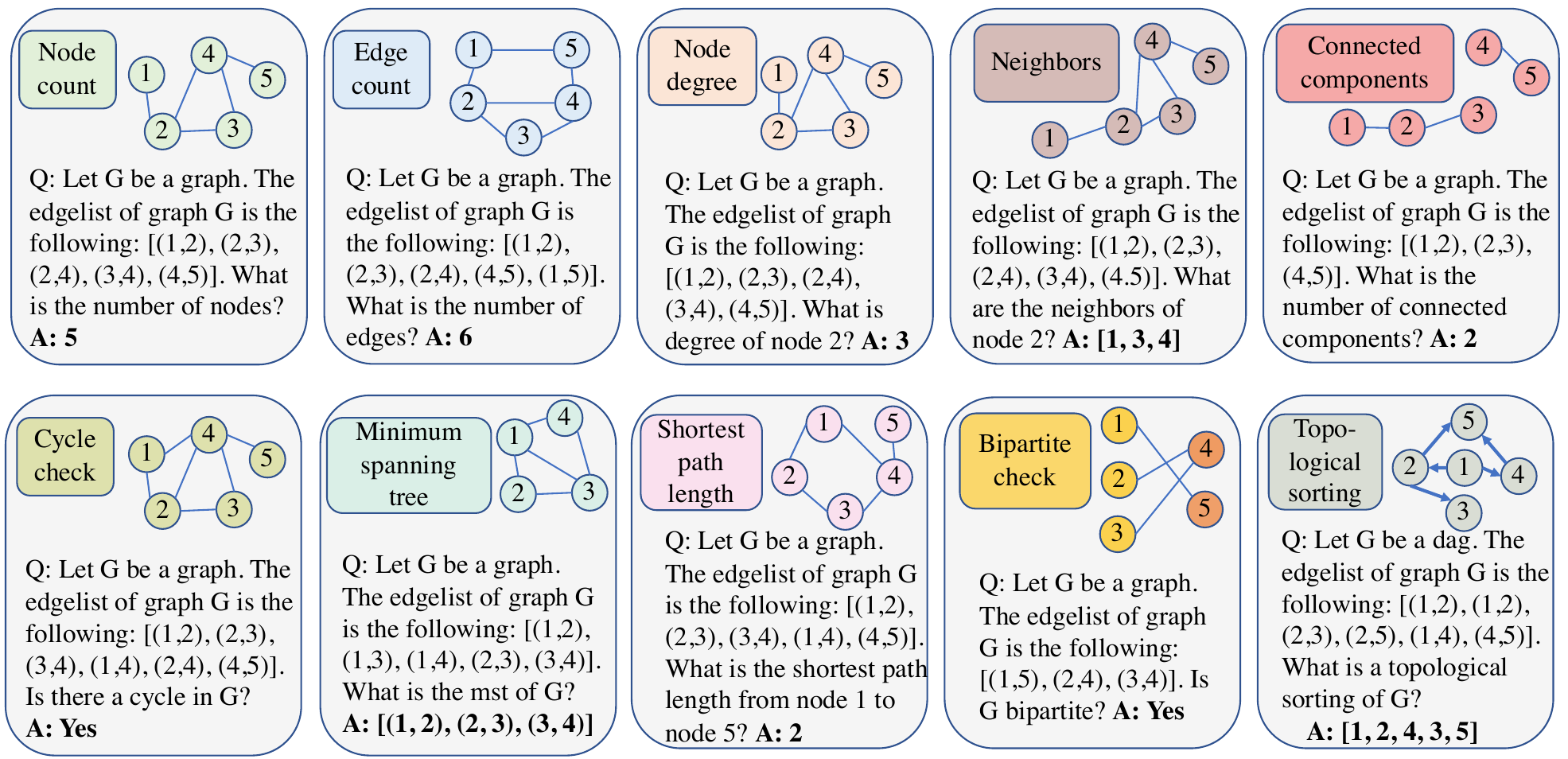}
    \caption{The proposed graph dataset.}
    \label{fig:dataset}
\end{figure*}

\paragraph{Graph tasks.}
There exist many decision and optimization problems on graphs.
Several of those problems are hard to solve (\eg finding a clique with the largest possible number of nodes is known to be an NP-hard problem).
We cannot expect an LLM to be able to solve such problems in a short amount of time even when the input graphs are relatively small.
Thus, here we focus on problems that can be solved in polynomial time in the worst case by some graph algorithm.
We list below the $10$ considered graph problems.
\begin{enumerate}[leftmargin=1em, noitemsep, topsep=0pt]
	\item \textbf{Node count} - Count the number of nodes.
	\item \textbf{Edge count} - Count the number of edges.
	\item \textbf{Node degree} - Calculate the degree of a node.
	\item \textbf{Neighbors} - Find all nodes that are adjacent to a given node.
	\item \textbf{Connected components} - Count the number of connected components.
	\item \textbf{Cycle check} - Check if a graph contains a cycle.
	\item \textbf{Minimum spanning tree (MST)} - Find the minimum cardinality subset of edges of a given graph that connects all the vertices together, without any cycles.
	\item \textbf{Shortest path} - Calculate the shortest path length between two nodes in a graph.
	\item \textbf{Bipartite check} - Check if a graph is bipartite.
	\item \textbf{Topological sorting} - Calculate a linear ordering of the nodes of a given directed acyclic graph such that for every directed edge $(u,v)$ from node $u$ to node $v$, $u$ comes before $v$ in the ordering. 
\end{enumerate}
Note that some tasks are easier, while others are more complex.
For example, given the list of edges of the graph, \textit{Edge count} requires just counting the number of elements of the list, while \textit{Shortest path} is a more complex task since it generally requires further algorithmic steps to be performed to reach the solution.

\paragraph{Generated graphs.}
Even though the provided source code allows one to generate different types of graphs (\eg Erd{\H{o}}s--R{\'e}nyi graphs, Barab{\'a}si--Albert graphs, star graphs, etc.), in this study, due to monetary costs, we focus mainly on Erd{\H{o}}s--R{\'e}nyi graphs.
Therefore, for all tasks except \textit{Bipartite check} and \textit{Topological sorting}, the graphs about which the LLM is asked to reason are Erd{\H{o}}s--R{\'e}nyi graphs.
To construct such a graph, we need to choose the number of nodes $n$ and the edge probability $p$.
As will be discussed later, we construct datasets of varying complexity, and the value of $n$ depends on the type of the dataset.
Hyperparameter $p$ is sampled from $[0,1]$ with uniform probability.
For the \textit{Topological sorting} task, we construct Erd{\H{o}}s--R{\'e}nyi graphs and we transform them into directed acyclic graphs.
This is achieved by first mapping the nodes to integers, \ie $\{ 1,\ldots,n \}$, and then assigning direction to all edges such that they point from lower nodes to higher nodes.
Finally, for \textit{Bipartite check}, we either construct Erd{\H{o}}s--R{\'e}nyi graphs or random bipartite graphs.
To construct a random bipartite graph, we create two sets of nodes such that no set is empty and such that the sum of their cardinalities is $n$, and then edges between nodes of one set and nodes of the other are included in the graph with probability $p$ (where $p$ is sampled from $[0,1]$).

\paragraph{Generated problems.}
For each task, we construct three different datasets.
The difference between those datasets lies in the number of nodes of the produced graphs.
One dataset consists of small graphs, one consists of medium-sized graphs, and the last one consists of large graphs.
We denote those three datasets by S, M, and L, respectively.
The number of nodes of the graphs contained in those three datasets range between $5$ and $11$ nodes for S, $11$ and $21$ nodes for M, and $21$ and $51$ nodes for L.
The different tasks do not share the same datasets of graphs.
A different dataset is constructed for each task.
Note that dataset L consists of graphs significantly larger than the ones considered in prior work (\ie all graphs had $5$ and $35$ nodes in~\cite{wang2023can} and between $5$ and $20$ nodes in~\cite{fatemi2024talk}).
Our results thus also provide insights into the capabiblity of LLMs to perform reasoning tasks on \textit{larger graphs} than the ones considered in previous studies.
Once the graphs are generated, we create the prompts and we add to them pseudo-code instructions.
We have created such instructions for all $10$ considered tasks.
An overview of the proposed graph reasoning tasks is shown in Figure \ref{fig:dataset}.

\begin{table}

\resizebox{\columnwidth}{!}{%
\begin{tabular}{ ccccccc } 
 \toprule
  & \multirowcell{2}{\diagbox{Methods}{Tasks}} & \multirowcell{2}{Node\\count} & \multirowcell{2}{Edge\\count} & \multirowcell{2}{Node\\degree} & \multirowcell{2}{Neighbors} \\
  & \\
\midrule
\multirowcell{6}{S} & {\scshape 0-shot} & 99 & 78 & \bf 75 & \bf 90 \\ 
& {\scshape 1-shot} & \bf 100 & 76 & 72 & 67 \\
& {\scshape BaG} & 67 & 57 & 73 & 78\\
& {\scshape 0-CoT} & 82 & 67 & 70 & 77  \\
& {\scshape Pseudo} & 87 & \bf 90 & 56 & 75 \\ 
& {\scshape Pseudo+1-shot} & 95 & 82 & 60 & 68 \\
\midrule
\multirowcell{6}{M} & {\scshape 0-shot} & 88 & 16 & 24 & 42\\ 
& {\scshape 1-shot} & \bf 100 & 22 & 28 & 29\\
& {\scshape BaG} & 50 & 11 & 31 & 44 \\
& {\scshape 0-CoT} & 62 & 13 & \bf 46 & \bf 51  \\
& {\scshape Pseudo} & 79 & \bf 34 & 18 & 37 \\
& {\scshape Pseudo+1-shot} & 63 & 18 & 43 & 30  \\
\midrule
\multirowcell{6}{L} & {\scshape 0-shot} & \bf 100 & 2 & 6 & 12 \\ 
& {\scshape 1-shot} & 96 & 1 & 0 & 9\\
& {\scshape BaG} & 72 & 0 & 7 & \bf 13 \\
& {\scshape 0-CoT} & 7 & 2 & \bf 13 & 12 \\
& {\scshape Pseudo} & 62 & \bf 9 & 6 & \bf 13 \\ 
& {\scshape Pseudo+1-shot} & 20 & 2 & \bf 13 & \bf 13 \\
\bottomrule
\end{tabular}
}
\caption{Model GPT-3.5-Turbo-0125 results on simple tasks. Bold indicates best results.}
\label{gpt3.5-results1}
\end{table}

Note that besides \textit{Node degree}, \textit{Neighbors} and \textit{Shortest path}, the rest of the tasks correspond to graph-level properties.
For each one of those seven tasks and for each graph size (\ie S, M or L), we construct $100$ problems.
This gives rise to $2,100$ problems in total.
The \textit{Node degree} and \textit{Neighbors} tasks capture node-level properties of graphs.
For those tasks and for each graph size, we create $100$ graphs and from each one of those graphs, we randomly choose $5$ nodes to create problems.
This leads to $3,000$ more problems.
Finally, the \textit{Shortest path} task is defined between pairs of nodes.
Once again, for each graph size, we create $100$ graphs and from each one of those graphs, we randomly choose $5$ pairs of nodes that both belong to the same connected component to create problems.
This results into $1,500$ more problems.
Overall, our dataset contains $6,600$ problems.

\section{Experiments}

\paragraph{Baselines.}
We compare the proposed method against the following three prompting approaches: (1) zero-shot prompting ({\scshape 0-shot}); (2) one-shot in-context learning ({\scshape 1-shot})~\cite{brown2020language}; (3) Build-a-Graph prompting ({\scshape BaG})~\cite{wang2023can}; and (4) zero-shot chain-of-thought ({\scshape 0-CoT})~\cite{kojima2022large}.
{\scshape 0-shot} constructs a prompt that describes the task and asks the LLM to solve the task, without any prior training on the task. 
Besides just a description of the task, {\scshape 1-shot} also provides the model with one example of the task, along with the desired output.
{\scshape BaG} adds the sentence ``Let's construct a graph with the nodes and edges first'' to the task description. 
Last, {\scshape 0-CoT} adds the sentence ``Let's think step by step'' to the task description to let the model generate its own Chain-of-Thoughts.

\begin{table*}[t]
\centering
\resizebox{0.76\textwidth}{!}{%
\begin{tabular}{ cccccccc } 
 \toprule
  & \multirowcell{2}{\diagbox{Methods}{Tasks}} & \multirowcell{2}{Connected\\components} & \multirowcell{2}{Cycle\\check} & \multirowcell{2}{MST} & \multirowcell{2}{Shortest\\path} & \multirowcell{2}{Bipartite\\check} & \multirowcell{2}{Topological\\sorting} \\ 
  & \\
\midrule
\multirowcell{6}{S} & {\scshape 0-shot} & 45 & 43 & 61 & 42 & 31 & \bf 88 \\ 
& {\scshape 1-shot} & \bf 86 & 44 & 47 & \bf 73 & \bf 61 & 77  \\
& {\scshape BaG} & 4 & 23 & 19 & 18 & 48 & \bf 88\\
& {\scshape 0-CoT} & 30 & 47 & 16 & 25 & 51 & 62 \\
& {\scshape Pseudo} & 76 & 76 & 61 & 50 & 52 & 72 \\ 
& {\scshape Pseudo+1-shot} & 69 & \bf 79 & \bf 64 & 59 & \bf 61 & 81 \\
\midrule
\multirowcell{6}{M} & {\scshape 0-shot} & 57 & 7 & 23 & 15 & \bf 51 & \bf 59 \\ 
& {\scshape 1-shot} & \bf 91 & 46 & 6 & \bf 61 & \bf 51 & 33 \\
& {\scshape BaG} & 3 & 8 & 7 & 8 & 26 & 34\\
& {\scshape 0-CoT} & 2 & 39 & 0 & 7 & 45 & 25 \\
& {\scshape Pseudo} & 66 & \bf 47 & 17 & 35 & 48 & 55 \\
& {\scshape Pseudo+1-shot} & 47 & \bf 47 & \bf 27 & 51 & 42 & 36 \\
\midrule
\multirowcell{6}{L} & {\scshape 0-shot} & \bf 85 & 34 & 2 & 7 & 43 & \bf 28 \\ 
& {\scshape 1-shot} & 40 & 21 & 1 & 27 & 42 & 13 \\
& {\scshape BaG} & 2 & 1 & 4 & 14 & 17 & 8\\
& {\scshape 0-CoT} & 0 & 6 & 0 & 2 & 48 & 6 \\
& {\scshape Pseudo} & 49 & \bf 71 & 10 & 22 & \bf 49 & 14 \\ 
& {\scshape Pseudo+1-shot} & 22 & 23 & \bf 27 & \bf 34 & 31 & 9\\
\bottomrule
\end{tabular}
}
\caption{Model GPT-3.5-Turbo-0125 results on the complex graph reasoning tasks. Results present accuracy in percentage (\%). Bold indicates best results.}
\label{gpt3.5-results2}
\end{table*}

\paragraph{Models and Settings.}
We evaluate two popular LLMs, namely GPT-3.5-Turbo and Mixtral 7x8B, thus representing both proprietary and open source LLMs.
For all baselines we set the parameter temperature = 0 in order to make results more deterministic and avoid randomness.
As discussed above, we evaluate LLMs and various prompting techniques mainly on Erd{\H{o}}s--R{\'e}nyi graphs due to monetary costs, while we plan to evaluate the proposed method on other types of graphs in the future.
We use two different variants of the proposed method.
In the first variant ({\scshape Pseudo}), we provide the LLM with the task description and the pseudocode to solve it, while in the second variant ({\scshape Pseudo + \scshape 1-shot}), we also provide the model with one example of the task, along with the desired output. 
Previous works have found that graph encoding functions (\ie how to represent the graph in natural language) have a significant impact on the performance of LLMs in the different graph tasks~\cite{guo2023gpt4graph,fatemi2024talk}.
In this paper, we choose to represent each graph by its list of edges since it was shown that it outperforms other common representations~\cite{guo2023gpt4graph}.

\paragraph{Evaluation metric.}
In all considered tasks, we are interested in finding whether the LLM provides the correct answer to a given query.
We thus measure performance by computing the accuracy, \ie correct answers/total queries.



\paragraph{Performance on graph tasks.}
We first split the $10$ different graph reasoning tasks into simpler tasks and more complex tasks.
In the first part of our analysis, we focus on the simple tasks (\ie \textit{Node count}, \textit{Edge count}, \textit{Node degree} and \textit{Neighbors}).
We evaluate the different prompting approaches and initially employ GPT-3.5-Turbo-0125 as our LLM.
Table~\ref{gpt3.5-results1} illustrates the results for these experiments.
We observe that {\scshape 0-shot} and {\scshape 1-shot} prompting can accurately count the number of nodes of a graph even if the graph is large.
Quite surprisingly, pseudo-code prompting fails to achieve similar levels of performance in this task.
However, {\scshape Pseudo} is the best-performing method in the \textit{Edge count} task.
In the \textit{Node degree} and \textit{Neighbors} tasks, no method outperforms consistently all the other methods.
For small graphs, the LLM correctly answers more than half of the queries no matter the prompting technique.
Besides the \textit{Neighbors} task, {\scshape 0-CoT} generally does not lead to improvements.
As expected, the performance of the model decreases as the size of graphs increases.
Overall, we observe that when the size of graphs is small, GPT3.5 performs quite well in the $4$ simple reasoning tasks even when no examples or assistance is provided.

We next evaluate the GPT-3.5 model in the remaining $6$ tasks (\ie \textit{Connected components}, \textit{Cycle check}, \textit{MST}, \textit{Shortest path}, \textit{Bipartite check} and \textit{Topological sorting}).
The results for these experiments are shown in Table~\ref{gpt3.5-results2}.
While one would expect the {\scshape 0-shot} approach to fail in all these tasks, we observe that it excels in \textit{Topological sorting}.
The example that the {\scshape 1-shot} method provides to the LLM seems to have a significant impact in some tasks, such as in identifying connected components and in computing shortest path distances. The {\scshape 0-CoT} method is the worst-performing prompting technique, likely due to its inability to generate the actual reasoning steps needed to solve the problem.
Incorporating pseudo-code into the prompt yields considerable improvements in some tasks, such as in computing shortest path lengths and in checking whether graph contain cycles where it provides the highest accuracy.
The {\scshape Pseudo+1-shot} approach is the best-performing prompting technique in the \textit{MST} task and in computing shortest path lengths in large graphs.
Surprisingly, in the \textit{Connected components} and \textit{Bipartite check} tasks, the size of the graphs does not seem to have any impact on the performance of the GPT-3.5 model.


\begin{table}[t]
\resizebox{\columnwidth}{!}{%
\begin{tabular}{ ccccccccc } 
 \toprule
  & \multirowcell{2}{\diagbox{Methods}{Tasks}} & \multirowcell{2}{Node\\count} & \multirowcell{2}{Edge\\count} & \multirowcell{2}{Node\\degree} & \multirowcell{2}{Neighbors}  \\ 
  & \\
\midrule
\multirow{6}{*}{S} & {\scshape 0-shot} & 92 & 56 & 56 & 65 \\ 
& {\scshape 1-shot} & 90 & 31 & 39 & 68 \\
& {\scshape BaG} & 92 & 51 & 64 & 60\\
& {\scshape 0-CoT} & 88 & 42 & 70 & \bf 75 \\
& {\scshape Pseudo} & 89 & 83 & 63 & 63 \\ 
& {\scshape Pseudo+1-shot} & \bf 97 & \bf 99 & \bf 73 & 64 \\
\midrule
\multirow{6}{*}{M} & {\scshape 0-shot} & 89 & 8 & 23 & 27 \\ 
& {\scshape 1-shot} & 88 & 9 & 7 & 31 \\
& {\scshape BaG} & 92 & 9 & 29 & 27 \\
& {\scshape 0-CoT} & \bf 93 & 3 & \bf 34 & \bf 37 \\
& {\scshape Pseudo} & 84 & 29 & 27 & 28 \\ 
& {\scshape Pseudo+1-shot} & 81 & \bf 89 & 31 & 27 \\
\midrule
\multirow{6}{*}{L} & {\scshape 0-shot} & 65 & 1 & 9 & 7\\ 
& {\scshape 1-shot} & 86 & 0 & 1 & 8 \\
& {\scshape BaG} & \bf 90 & 0 & \bf 12 & 7\\
& {\scshape 0-CoT} & 83 & 2 & 11 & \bf 10 \\
& {\scshape Pseudo} & 80 & 7 & 7 & 7 \\ 
& {\scshape Pseudo+1-shot} & 56 & \bf 14 & 8 & 5\\
\bottomrule
\end{tabular}
}
\caption{Mixtral results on simple tasks. Results present accuracy in percentage (\%). Bold indicates best results.}
\label{mixtral1}
\end{table}

We also experiment with the open-source Mixtral 7x8B model.
The obtained results for the simpler tasks are shown in Figure~\ref{mixtral1}.
We observe that no matter what prompting method we use, the model can always quite accurately count the number of nodes of the input graphs.
However, in the rest of the tasks, {\scshape 0-shot} and {\scshape 1-shot} fail to achieve high levels of accuracy, especially for medium-sized and large graphs.
In the \textit{Edge count} task, these methods return a correct answer for less than $10\%$ of the queries when the input graphs are not small.
The results also suggest that {\scshape 0-CoT} and {\scshape BaG} lead to performance improvements in most cases.
Pseudo-code prompting also leads to significant performance gains in most cases.
For example, {\scshape Pseudo+1-shot} achieves the highest accuracy in the \textit{Node count}, \textit{Edge count}, and \textit{Node degree} tasks, thus demonstrating how useful pseudo-code prompting is for less powerful LLMs.
Specifically, in the \textit{Edge count} and \textit{Node degree} tasks and for small graphs, {\scshape Pseudo+1-shot} led to a respective relative increase of $76.8\%$ and $30.4\%$ in accuracy over {\scshape 0-shot}.
Furthermore, in the \textit{Edge count} task and for medium-sized graphs, {\scshape Pseudo+1-shot} resulted in an impressive relative increase of $1012.5\%$ in accuracy.
Finally, we should note that in most tasks, Mixtral's performance also decreases as the size of graphs increases.

\begin{table*}[t]
\centering
\resizebox{0.76\textwidth}{!}{%
\begin{tabular}{ ccccccccc } 
 \toprule
& \multirowcell{2}{\diagbox{Methods}{Tasks}} & \multirowcell{2}{Connected\\components} & \multirowcell{2}{Cycle\\check} & \multirowcell{2}{MST} & \multirowcell{2}{Shortest\\ path} & \multirowcell{2}{Bipartite\\check} & \multirowcell{2}{Topological\\sorting} \\ 
  & \\
\midrule
\multirow{6}{*}{S} & {\scshape 0-shot} & 35 & 85 & 24 & 48 & 44 & 39 \\ 
& {\scshape 1-shot} & 47 & 77 & 18 & 52 & 47 & \bf 57 \\
& {\scshape BaG} & \bf 78 & 82 & 19 & 28 & 39 & 51\\
& {\scshape 0-CoT} & 70 & \bf 90 & 27 & 55 & \bf 58 & 35 \\
& {\scshape Pseudo} & 62 & 33 & 24 & 50 & 47 & 40 \\ 
& {\scshape Pseudo+1-shot} & 75 & 51 & \bf 42 & \bf 56 & 53 & 47 \\
\midrule
\multirow{6}{*}{M} & {\scshape 0-shot} & 40 & \bf 93 & 6 & 30 & 42 & 6 \\ 
& {\scshape 1-shot} & 31 & 75 & 7 & \bf 50 & 50 & 11 \\
& {\scshape BaG} & \bf 65 & 86 & \bf 8 & 28 & \bf 53 & 9 \\
& {\scshape 0-CoT} & 57 & 90 & \bf 8 & 35 & 42 & 8 \\
& {\scshape Pseudo} & 63 & 36 & 5 & 27 & 47 & 12 \\ 
& {\scshape Pseudo+1-shot} & 42 & 40 & 5 & 40 & 48 & \bf 18 \\
\midrule
\multirow{6}{*}{L} & {\scshape0-shot} & 34 & 86 & 1 & 17 & 51 & 4 \\ 
& {\scshape 1-shot} & 29 & 69 & 1 & \bf 25 & 48 & \bf 11 \\
& {\scshape BaG} & 25 & 77 & 1 & 10 & 45 & 2\\
& {\scshape 0-CoT} & 27 & \bf 92 & 1 & 15 & \bf 53 & 2 \\
& {\scshape Pseudo} & \bf 41 & 31 & 1 & 13 & 44 & 5 \\ 
& {\scshape Pseudo+1-shot} & 18 & 35 & 1 & 24 & 41 & 10\\
\bottomrule
\end{tabular}
}
\caption{Mixtral 8x7B results on the complex graph reasoning tasks. The results present accuracy in percentage (\%). Bold indicates best results.}
\label{mixtral2}
\end{table*}

The results for the more complex graph reasoning tasks are illustrated in Table~\ref{mixtral2}.
We observe that when pseudo-code is added to the prompt, it becomes harder for Mixtral to detect whether the input graph contains any cycle.
However, the use of pseudo-code proves crucial for some other tasks such as \textit{Connected components}.
Interestingly, for small graphs, the {\scshape Pseudo+1-shot} approach results in a relative increase of $114.3\%$, $75\%$ and $16.7\%$ in accuracy over {\scshape 0-shot} in the \textit{Connected components}, \textit{MST} and \textit{Shortest path} tasks, respectively.
Likewise, for medium-sized graphs, the use of pseudo-code use leads to a relative increase of $57.5\%$ in accuracy over {\scshape 0-shot} in \textit{Connected components}.
These findings clearly indicate that augmenting the prompt with pseudo-code instructions and corresponding examples can significantly enhance accuracy in both simple and complex graph reasoning tasks.


\begin{wraptable}{r}{4cm}
\centering
\resizebox{0.5\columnwidth}{!}{%
\footnotesize
\begin{tabular}{ cccc } 
 \toprule
& 1 &  2 & 3\\ 
\midrule
MST & \bf 31 & 24 & 29 \\
Neighbors & 40 & \bf 63 & 46 \\
\bottomrule
\end{tabular}
}
\caption{Results with different pseudo-code styles.}
\label{pseudo-code}
\end{wraptable}
\paragraph{Pseudo-code style.}
We next investigated what is the impact of the type of utilized pseudo-code on the performance of the LLM.
Table~\ref{pseudo-code} illustrates the results with different pseudo-code styles on Mixtral (1: Python, 2: Pseudo, 3: Complex).
\begin{figure}
\centering
    \includegraphics[width=0.8\columnwidth]{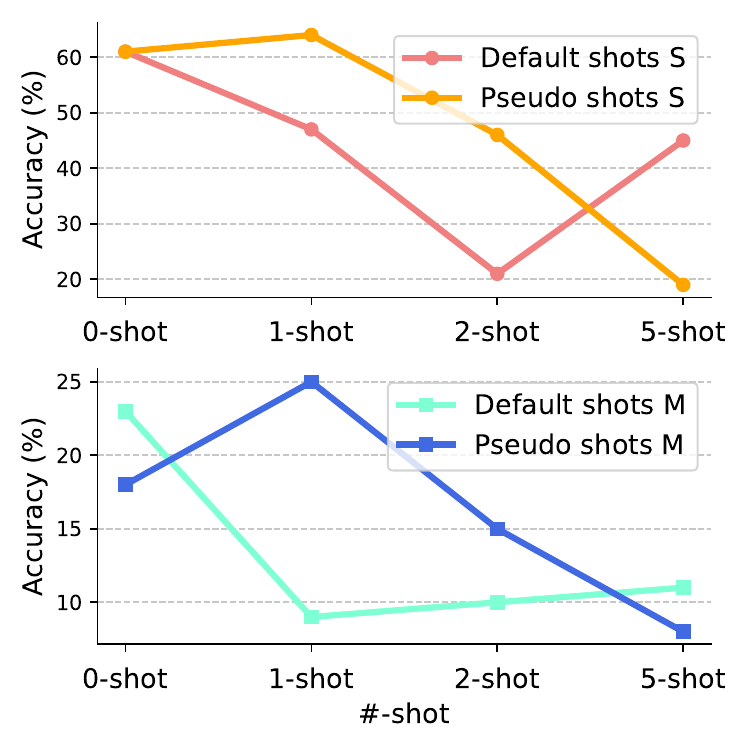}
  \caption{\#-shot results in minimum spanning tree.}
  \label{shots}
\end{figure}
We observe that the results are mixed.
Plain pseudo-code outperforms the rest in the \textit{Neighbors} task, while Python code achieves the highest accuracy in the \textit{MST} task.
The pseudo-code that consists of multiple functions instead of a single one is the second-best method in both tasks.
By examining the results, we observed that the LLM struggles when presented with nested loops and recursive functions.

\paragraph{One vs. few examples.}
We also investigated whether we can obtain performance gains by increasing the number of examples provided to the model.
Figure~\ref{shots} illustrates performance of GPT-3.5 in the small subset of the \textit{MSE} task as a function of the number of examples.
The results suggest that in case pseudo-code is present, a single example suffices.
Unlike the {\scshape 0-shot} method, where adding more examples enhances the reasoning abilities of the LLM, our approach does not seem to benefit from multiple examples. 
Therefore, the proposed method appears to be more cost-efficient that other prompting techniques, as one example is enough to lead to performance improvements.
Creating multiple examples, particularly in the context of graphs, can be time-consuming and resource-intensive. 

\paragraph{Summary.}


We next present our main findings:
\begin{itemize}[left=0pt, noitemsep, topsep=0pt]
\item \textbf{For most tasks, the size of the input graphs has a significant impact on the LLMs' performance.}
With the exception of \textit{Node count}, \textit{Connected components} and \textit{Bipartite check}, in all other tasks, performance decreases significantly as the size of the graphs increases.

\item \textbf{LLMs can count nodes, but they cannot count edges.}
While LLMs could quite accurately count the number of nodes of all graphs, no method achieved an accuracy greater than $14\%$ in counting the number of edges of large graphs.

\item \textbf{Pseudo-code is useful for tasks that LLMs struggle to solve.}
Pseudo-code offered significant improvements in the \textit{Edge count} and \textit{MST} tasks, where the failure rate of LLMs is high.

\item \textbf{There exist tasks where pseudo-code might improve the performance of one LLM, but lead another LLM to lower levels of performance.}
{\scshape Pseudo} significantly outperforms {\scshape 0-shot} in the \textit{Cycle check} task when using GPT-3.5.
On the other hand, {\scshape Pseudo} is significantly outperformed by {\scshape 0-shot} in the same task with Mixtral.

\item \textbf{Carefully designed prompting can improve the performance of LLMs.}
In almost all our experiments, {\scshape 0-shot} was outperformed by the rest of the prompting techniques.
This direction is computationally less demanding than fine-tuning pre-trained LLMs.

\item \textbf{In the presence of pseudo-code, a single example is enough.}
Even in complex graph reasoning tasks, prompting with pseudo-code does not need several examples to reach its full potential.



\end{itemize}

\section{Conclusion}
In this work, we explored whether prompting with pseudo-code instructions can enhance LLMs' reasoning on simple and complex graph tasks. 
Experiments with GPT-3.5 and Mixtral show that pseudo-code prompts improve performance across various graph tasks. 
However, performance declines as graph size increases. 
This highlights the need for further research on prompting techniques for large graphs. 
Our focus is on improving both reasoning and interpretability, showing LLMs can solve problems while making their reasoning steps explicit.

\section*{Limitations}
\paragraph{Pseudo-code prompts need to be carefully designed or might not be available.}
To get the most out of pseudo-code, careful design is needed.
Simple coding is preferred and complex structures such as nested loops or recursive functions should be avoided.
We assume that pseudo-code is either directly available or there is access to the technical expertise required to write it.

\paragraph{Evaluation} Automatically evaluating the performance of LLMs is by definition a hard task. In order to measure the performance, we search for the result in the LLM output. 
Therefore, some degree of ambiguity, variation in phrasing, and differences in reasoning approaches are inevitable. 
As a result, certain errors are expected when aligning the generated output with predefined answers or benchmarks."

\bibliography{coling_latex}

\begin{thebibliography}{34}
\providecommand{\natexlab}[1]{#1}

\bibitem[{Ammanabrolu and Riedl(2021)}]{ammanabrolu2021learning}
Prithviraj Ammanabrolu and Mark Riedl. 2021.
\newblock {Learning Knowledge Graph-based World Models of Textual
  Environments}.
\newblock In \emph{Advances in Neural Information Processing Systems}, pages
  3720--3731.

\bibitem[{Besta et~al.(2024)Besta, Blach, Kubicek, Gerstenberger, Podstawski,
  Gianinazzi, Gajda, Lehmann, Niewiadomski, Nyczyk et~al.}]{besta2024graph}
Maciej Besta, Nils Blach, Ales Kubicek, Robert Gerstenberger, Michal
  Podstawski, Lukas Gianinazzi, Joanna Gajda, Tomasz Lehmann, Hubert
  Niewiadomski, Piotr Nyczyk, et~al. 2024.
\newblock {Graph of Thoughts: Solving Elaborate Problems with Large Language
  Models}.
\newblock In \emph{Proceedings of the 38th AAAI Conference on Artificial
  Intelligence}, pages 17682--17690.

\bibitem[{Brown et~al.(2020)Brown, Mann, Ryder, Subbiah, Kaplan, Dhariwal,
  Neelakantan, Shyam, Sastry, Askell et~al.}]{brown2020language}
Tom Brown, Benjamin Mann, Nick Ryder, Melanie Subbiah, Jared~D Kaplan, Prafulla
  Dhariwal, Arvind Neelakantan, Pranav Shyam, Girish Sastry, Amanda Askell,
  et~al. 2020.
\newblock {Language Models are Few-Shot Learners}.
\newblock In \emph{Advances in Neural Information Processing Systems}, pages
  1877--1901.

\bibitem[{Bubeck et~al.(2023)Bubeck, Chandrasekaran, Eldan, Gehrke, Horvitz,
  Kamar, Lee, Lee, Li, Lundberg et~al.}]{bubeck2023sparks}
S{\'e}bastien Bubeck, Varun Chandrasekaran, Ronen Eldan, Johannes Gehrke, Eric
  Horvitz, Ece Kamar, Peter Lee, Yin~Tat Lee, Yuanzhi Li, Scott Lundberg,
  et~al. 2023.
\newblock {Sparks of Artificial General Intelligence: Early experiments with
  GPT-4}.
\newblock \emph{arXiv preprint arXiv:2303.12712}.

\bibitem[{Chen et~al.(2023)Chen, Ma, Wang, and Cohen}]{chen2023program}
Wenhu Chen, Xueguang Ma, Xinyi Wang, and William~W Cohen. 2023.
\newblock {Program of Thoughts Prompting: Disentangling Computation from
  Reasoning for Numerical Reasoning Tasks}.
\newblock \emph{Transactions on Machine Learning Research}.

\bibitem[{Chen et~al.(2024)Chen, Mao, Li, Jin, Wen, Wei, Wang, Yin, Fan, Liu
  et~al.}]{chen2024exploring}
Zhikai Chen, Haitao Mao, Hang Li, Wei Jin, Hongzhi Wen, Xiaochi Wei, Shuaiqiang
  Wang, Dawei Yin, Wenqi Fan, Hui Liu, et~al. 2024.
\newblock {Exploring the Potential of Large Language Models (LLMs) in Learning
  on Graphs}.
\newblock \emph{ACM SIGKDD Explorations Newsletter}, 25(2):42--61.

\bibitem[{Devlin et~al.(2019)Devlin, Ming-Wei, Lee, and
  Toutanova}]{devlin2019bert}
Jacob Devlin, Chang Ming-Wei, Kenton Lee, and Kristina Toutanova. 2019.
\newblock {BERT: Pre-training of Deep Bidirectional Transformers for Language
  Understanding}.
\newblock In \emph{Proceedings of NAACL-HLT}, pages 4171--4186.

\bibitem[{Duan et~al.(2023)Duan, Liu, Chua, Yan, Ooi, Xie, and
  He}]{duan2023simteg}
Keyu Duan, Qian Liu, Tat-Seng Chua, Shuicheng Yan, Wei~Tsang Ooi, Qizhe Xie,
  and Junxian He. 2023.
\newblock {SimTeG: A Frustratingly Simple Approach Improves Textual Graph
  Learning}.
\newblock \emph{arXiv preprint arXiv:2308.02565}.

\bibitem[{Fatemi et~al.(2024)Fatemi, Halcrow, and Perozzi}]{fatemi2024talk}
Bahare Fatemi, Jonathan Halcrow, and Bryan Perozzi. 2024.
\newblock {Talk like a Graph: Encoding Graphs for Large Language Models}.
\newblock In \emph{The 12th International Conference on Learning
  Representations}.

\bibitem[{Guan et~al.(2024)Guan, Liu, Lin, Lu, He, Han, and
  Sun}]{guan2024mitigating}
Xinyan Guan, Yanjiang Liu, Hongyu Lin, Yaojie Lu, Ben He, Xianpei Han, and
  Le~Sun. 2024.
\newblock {Mitigating Large Language Model Hallucinations via Autonomous
  Knowledge Graph-Based Retrofitting}.
\newblock In \emph{Proceedings of the 38th AAAI Conference on Artificial
  Intelligence}, pages 18126--18134.

\bibitem[{Guo et~al.(2023)Guo, Du, and Liu}]{guo2023gpt4graph}
Jiayan Guo, Lun Du, and Hengyu Liu. 2023.
\newblock {GPT4Graph: Can Large Language Models Understand Graph Structured
  Data? An Empirical Evaluation and Benchmarking}.
\newblock \emph{arXiv preprint arXiv:2305.15066}.

\bibitem[{Guo et~al.(2024)Guo, Xia, Yu, Wang, Yang, Wei, Pang, Chua, and
  Huang}]{guo2024graphedit}
Zirui Guo, Lianghao Xia, Yanhua Yu, Yuling Wang, Zixuan Yang, Wei Wei, Liang
  Pang, Tat-Seng Chua, and Chao Huang. 2024.
\newblock {GraphEdit: Large Language Models for Graph Structure Learning}.
\newblock \emph{arXiv preprint arXiv:2402.15183}.

\bibitem[{He et~al.(2024)He, Bresson, Laurent, Perold, LeCun, and
  Hooi}]{he2024harnessing}
Xiaoxin He, Xavier Bresson, Thomas Laurent, Adam Perold, Yann LeCun, and Bryan
  Hooi. 2024.
\newblock {Harnessing Explanations: LLM-to-LM Interpreter for Enhanced
  Text-Attributed Graph Representation Learning}.
\newblock In \emph{The 12th International Conference on Learning
  Representations}.

\bibitem[{Hu et~al.(2023)Hu, Zhang, and Zhao}]{hu2023beyond}
Yuntong Hu, Zheng Zhang, and Liang Zhao. 2023.
\newblock {Beyond Text: A Deep Dive into Large Language Models' Ability on
  Understanding Graph Data}.
\newblock In \emph{NeurIPS 2023 Workshop: New Frontiers in Graph Learning}.

\bibitem[{Jung et~al.(2022)Jung, Qin, Welleck, Brahman, Bhagavatula, Le~Bras,
  and Choi}]{jung2022maieutic}
Jaehun Jung, Lianhui Qin, Sean Welleck, Faeze Brahman, Chandra Bhagavatula,
  Ronan Le~Bras, and Yejin Choi. 2022.
\newblock {Maieutic Prompting: Logically Consistent Reasoning with Recursive
  Explanations}.
\newblock In \emph{Proceedings of the 2022 Conference on Empirical Methods in
  Natural Language Processing}, pages 1266--1279.

\bibitem[{Kim et~al.(2022)Kim, Nguyen, Min, Cho, Lee, Lee, and
  Hong}]{kim2022pure}
Jinwoo Kim, Dat Nguyen, Seonwoo Min, Sungjun Cho, Moontae Lee, Honglak Lee, and
  Seunghoon Hong. 2022.
\newblock {Pure Transformers are Powerful Graph Learners}.
\newblock In \emph{Advances in Neural Information Processing Systems}, pages
  14582--14595.

\bibitem[{Kojima et~al.(2022)Kojima, Gu, Reid, Matsuo, and
  Iwasawa}]{kojima2022large}
Takeshi Kojima, Shixiang~Shane Gu, Machel Reid, Yutaka Matsuo, and Yusuke
  Iwasawa. 2022.
\newblock {Large Language Models are Zero-Shot Reasoners}.
\newblock pages 22199--22213.

\bibitem[{Liu et~al.(2023)Liu, Yuan, Fu, Jiang, Hayashi, and
  Neubig}]{liu2023pre}
Pengfei Liu, Weizhe Yuan, Jinlan Fu, Zhengbao Jiang, Hiroaki Hayashi, and
  Graham Neubig. 2023.
\newblock {Pre-train, Prompt, and Predict: A Systematic Survey of Prompting
  Methods in Natural Language Processing}.
\newblock \emph{ACM Computing Surveys}, 55(9):1--35.

\bibitem[{Luo et~al.(2024)Luo, Li, Haffari, and Pan}]{luo2024reasoning}
Linhao Luo, Yuan-Fang Li, Gholamreza Haffari, and Shirui Pan. 2024.
\newblock {Reasoning on Graphs: Faithful and Interpretable Large Language Model
  Reasoning}.
\newblock In \emph{The 12th International Conference on Learning
  Representations}.

\bibitem[{Mishra et~al.(2023)Mishra, Kumar, Bhat, Murthy, Contractor, and
  Tamilselvam}]{mishra2023prompting}
Mayank Mishra, Prince Kumar, Riyaz Bhat, Rudra Murthy, Danish Contractor, and
  Srikanth Tamilselvam. 2023.
\newblock {Prompting with Pseudo-Code Instructions}.
\newblock In \emph{Proceedings of the 2023 Conference on Empirical Methods in
  Natural Language Processing}, pages 15178--15197.

\bibitem[{Ouyang et~al.(2022)Ouyang, Wu, Jiang, Almeida, Wainwright, Mishkin,
  Zhang, Agarwal, Slama, Ray et~al.}]{ouyang2022training}
Long Ouyang, Jeffrey Wu, Xu~Jiang, Diogo Almeida, Carroll Wainwright, Pamela
  Mishkin, Chong Zhang, Sandhini Agarwal, Katarina Slama, Alex Ray, et~al.
  2022.
\newblock Training language models to follow instructions with human feedback.
\newblock In \emph{Advances in Neural Information Processing Systems}, pages
  27730--27744.

\bibitem[{Pan et~al.(2024)Pan, Luo, Wang, Chen, Wang, and Wu}]{pan2024unifying}
Shirui Pan, Linhao Luo, Yufei Wang, Chen Chen, Jiapu Wang, and Xindong Wu.
  2024.
\newblock {Unifying Large Language Models and Knowledge Graphs: A Roadmap}.
\newblock \emph{IEEE Transactions on Knowledge and Data Engineering}.

\bibitem[{Perozzi et~al.(2024)Perozzi, Fatemi, Zelle, Tsitsulin, Kazemi,
  Al-Rfou, and Halcrow}]{perozzi2024let}
Bryan Perozzi, Bahare Fatemi, Dustin Zelle, Anton Tsitsulin, Mehran Kazemi,
  Rami Al-Rfou, and Jonathan Halcrow. 2024.
\newblock {Let Your Graph Do the Talking: Encoding Structured Data for LLMs}.
\newblock \emph{arXiv preprint arXiv:2402.05862}.

\bibitem[{Poesia et~al.(2022)Poesia, Polozov, Le, Tiwari, Soares, Meek, and
  Gulwani}]{poesia2022synchromesh}
Gabriel Poesia, Oleksandr Polozov, Vu~Le, Ashish Tiwari, Gustavo Soares,
  Christopher Meek, and Sumit Gulwani. 2022.
\newblock {Synchromesh: Reliable Code Generation from Pre-trained Language
  Models}.
\newblock In \emph{The 10th International Conference on Learning
  Representations}.

\bibitem[{Sun et~al.(2023)Sun, Ren, Ma, and Zhang}]{sun2023large}
Shengyin Sun, Yuxiang Ren, Chen Ma, and Xuecang Zhang. 2023.
\newblock {Large Language Models as Topological Structure Enhancers for
  Text-Attributed Graphs}.
\newblock \emph{arXiv preprint arXiv:2311.14324}.

\bibitem[{Thirunavukarasu et~al.(2023)Thirunavukarasu, Ting, Elangovan,
  Gutierrez, Tan, and Ting}]{thirunavukarasu2023large}
Arun~James Thirunavukarasu, Darren Shu~Jeng Ting, Kabilan Elangovan, Laura
  Gutierrez, Ting~Fang Tan, and Daniel Shu~Wei Ting. 2023.
\newblock Large language models in medicine.
\newblock \emph{Nature medicine}, 29(8):1930--1940.

\bibitem[{Vaswani et~al.(2017)Vaswani, Shazeer, Parmar, Uszkoreit, Jones,
  Gomez, Kaiser, and Polosukhin}]{vaswani2017attention}
Ashish Vaswani, Noam Shazeer, Niki Parmar, Jakob Uszkoreit, Llion Jones,
  Aidan~N Gomez, {\L}ukasz Kaiser, and Illia Polosukhin. 2017.
\newblock {Attention Is All You Need}.
\newblock In \emph{Advances in Neural Information Processing Systems}.

\bibitem[{Wang et~al.(2023{\natexlab{a}})Wang, Feng, He, Tan, Han, and
  Tsvetkov}]{wang2023can}
Heng Wang, Shangbin Feng, Tianxing He, Zhaoxuan Tan, Xiaochuang Han, and Yulia
  Tsvetkov. 2023{\natexlab{a}}.
\newblock {Can Language Models Solve Graph Problems in Natural Language?}
\newblock In \emph{Advances in Neural Information Processing Systems}.

\bibitem[{Wang et~al.(2023{\natexlab{b}})Wang, Wei, Schuurmans, Le, Chi,
  Narang, Chowdhery, and Zhou}]{wang2023self}
Xuezhi Wang, Jason Wei, Dale Schuurmans, Quoc~V Le, Ed~H Chi, Sharan Narang,
  Aakanksha Chowdhery, and Denny Zhou. 2023{\natexlab{b}}.
\newblock {Self-Consistency Improves Chain of Thought Reasoning in Language
  Models}.
\newblock In \emph{The 11th International Conference on Learning
  Representations}.

\bibitem[{Wei et~al.(2022)Wei, Wang, Schuurmans, Bosma, Xia, Chi, Le, Zhou
  et~al.}]{wei2022chain}
Jason Wei, Xuezhi Wang, Dale Schuurmans, Maarten Bosma, Fei Xia, Ed~Chi, Quoc~V
  Le, Denny Zhou, et~al. 2022.
\newblock {Chain-of-Thought Prompting Elicits Reasoning in Large Language
  Models}.
\newblock In \emph{Advances in Neural Information Processing Systems}, pages
  24824--24837.

\bibitem[{Yao et~al.(2023)Yao, Yu, Zhao, Shafran, Griffiths, Cao, and
  Narasimhan}]{yao2023tree}
Shunyu Yao, Dian Yu, Jeffrey Zhao, Izhak Shafran, Tom Griffiths, Yuan Cao, and
  Karthik Narasimhan. 2023.
\newblock {Tree of Thoughts: Deliberate Problem Solving with Large Language
  Models}.
\newblock In \emph{Advances in Neural Information Processing Systems}.

\bibitem[{Zhao et~al.(2023)Zhao, Zhuo, Shen, Qu, Liu, Bronstein, Zhu, and
  Tang}]{zhao2023graphtext}
Jianan Zhao, Le~Zhuo, Yikang Shen, Meng Qu, Kai Liu, Michael Bronstein,
  Zhaocheng Zhu, and Jian Tang. 2023.
\newblock {GraphText: Graph Reasoning in Text Space}.
\newblock \emph{arXiv preprint arXiv:2310.01089}.

\bibitem[{Zhou et~al.(2023)Zhou, Sch{\"a}rli, Hou, Wei, Scales, Wang,
  Schuurmans, Cui, Bousquet, Le et~al.}]{zhou2023least}
Denny Zhou, Nathanael Sch{\"a}rli, Le~Hou, Jason Wei, Nathan Scales, Xuezhi
  Wang, Dale Schuurmans, Claire Cui, Olivier Bousquet, Quoc~V Le, et~al. 2023.
\newblock {Least-to-Most Prompting Enables Complex Reasoning in Large Language
  Models}.
\newblock In \emph{The 11th International Conference on Learning
  Representations}.

\bibitem[{Zhou et~al.(2020)Zhou, Cui, Hu, Zhang, Yang, Liu, Wang, Li, and
  Sun}]{zhou2020graph}
Jie Zhou, Ganqu Cui, Shengding Hu, Zhengyan Zhang, Cheng Yang, Zhiyuan Liu,
  Lifeng Wang, Changcheng Li, and Maosong Sun. 2020.
\newblock Graph neural networks: A review of methods and applications.
\newblock \emph{AI Open}, 1:57--81.

\end{thebibliography}




\end{document}